\documentclass[runningheads]{llncs}

\usepackage{eccv}

\usepackage{eccvabbrv}

\usepackage{graphicx}
\usepackage{booktabs}
\usepackage{amsmath,amssymb}
\usepackage{color}
\usepackage{subcaption}
\usepackage{pifont}
\newcommand{\cmark}{\ding{51}}
\newcommand{\xmark}{\ding{55}}

\usepackage[accsupp]{axessibility}  

\usepackage{hyperref}

\usepackage[capitalize]{cleveref}

\usepackage{orcidlink}

\usepackage{algorithm}
\usepackage{algorithmic}

\begin{document}

\title{MoireMix: A Formula-Based Data Augmentation for Improving Image Classification Robustness} 
\titlerunning{MoireMix}

\author{Yuto Matsuo\inst{1,2} \and
Yoshihiro Fukuhara\inst{1,3} \and
Yuki M. Asano\inst{4} \and
Rintaro Yanagi\inst{1} \and
Hirokatsu Kataoka\inst{1,5} \and
Akio Nakamura\inst{2}}

\authorrunning{Y.~Matsuo \etal}

\institute{National Institute of Advanced Industrial Science and Technology (AIST)\\ \email{myis20fr@gmail.com} \and
Tokyo Denki University \and
Waseda University \and
Fundamental AI Lab, University of Technology Nuremberg \and
Visual Geometry Group, University of Oxford}

\maketitle

\begin{abstract}
Data augmentation is a key technique for improving the robustness of image classification models. However, many recent approaches rely on diffusion-based synthesis or complex feature mixing strategies, which introduce substantial computational overhead or require external datasets. In this work, we explore a different direction: procedural augmentation based on analytic interference patterns. Unlike conventional augmentation methods that rely on stochastic noise, feature mixing, or generative models, our approach exploits Moire interference to generate structured perturbations spanning a wide range of spatial frequencies.
We propose a lightweight augmentation method that procedurally generates Moire textures on-the-fly using a closed-form mathematical formulation. The patterns are synthesized directly in memory with negligible computational cost (0.0026 seconds per image), mixed with training images during training, and immediately discarded, enabling a storage-free augmentation pipeline without external data.
Extensive experiments with Vision Transformers demonstrate that the proposed method consistently improves robustness across multiple benchmarks, including ImageNet-C, ImageNet-R, and adversarial benchmarks, outperforming standard augmentation baselines and existing external-data-free augmentation approaches. These results suggest that analytic interference patterns provide a practical and efficient alternative to data-driven generative augmentation methods.

\end{abstract}

\section{Introduction}
\label{sec:intro}

\begin{figure}[t]
\centering
\includegraphics[width=\textwidth]{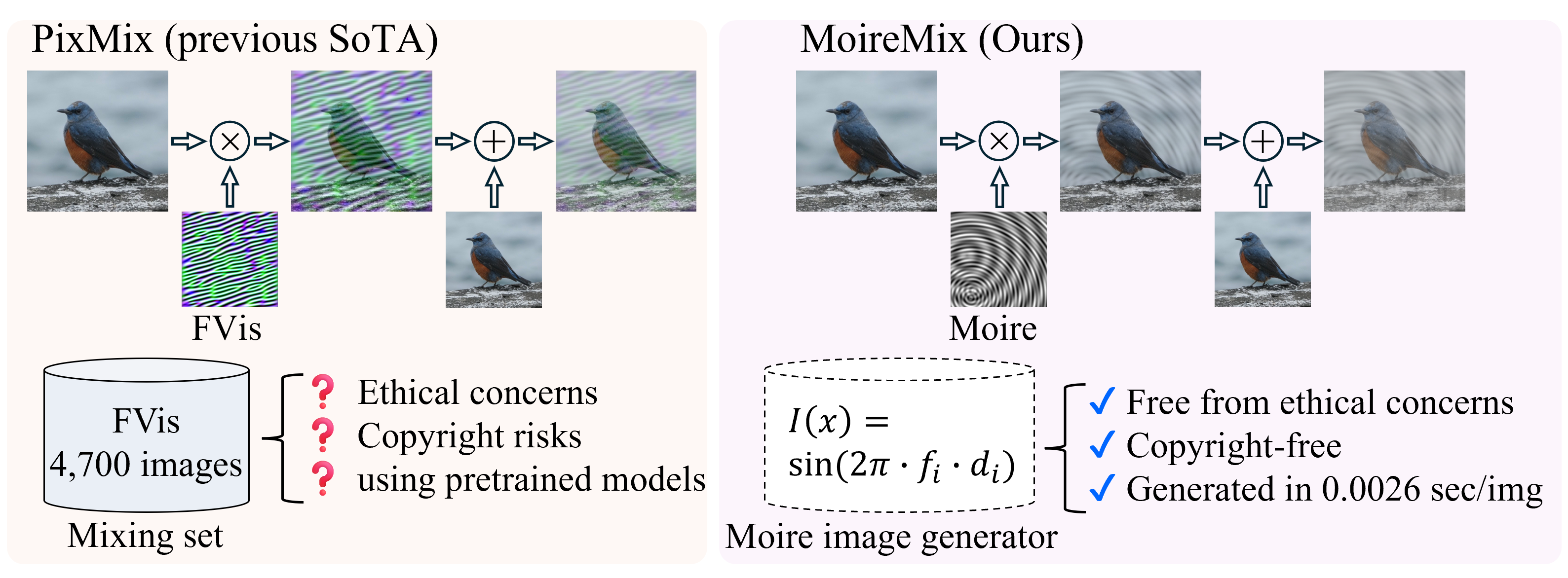}
\caption{Overview of our proposed MoireMix data augmentation framework. The method utilizes mathematical formulas to procedurally generate complex interference patterns on-the-fly during training.}
\label{fig:abst}
\end{figure}

Despite remarkable progress in image recognition, modern deep neural networks remain significantly less robust than human perception when exposed to common real-world visual corruptions such as noise, blur, and compression artifacts \cite{HendrycksCorruptions2019}. This lack of robustness poses a major obstacle to deploying deep learning systems in safety-critical and industrial environments, even though state-of-the-art architectures such as ResNet \cite{HeResNet2016}, Vision Transformers \cite{DosovitskiyViT2021}, and large-scale ImageNet models \cite{KrizhevskyImageNet2017} achieve near-human or even superhuman accuracy on clean benchmarks.

Data augmentation has emerged as one of the most effective strategies for improving robustness. Early approaches such as MixUp \cite{ZhangMixup2018} and CutMix \cite{YunCutMix2019} enhance generalization by mixing pixel intensities or spatial regions between training samples. More recent methods incorporate texture-based perturbations to expose models to a wider range of visual patterns. In particular, PixMix \cite{HendrycksPixMix2022} blends input images with diverse texture images to improve robustness while preserving semantic content. However, such approaches fundamentally rely on external collections of ``mixing images,'' which must be curated, stored, and maintained as secondary datasets.

A complementary line of research suggests that meaningful visual representations can emerge even from procedurally generated patterns rather than natural images. Formula-driven Supervised Learning (FDSL) \cite{KataokaIJCV2022} demonstrates that datasets synthesized from mathematical formulations, such as FractalDB and Dead Leaves \cite{BaradadNoise2021}, automatically generated contours \cite{KataokaRCDB2022}, sinusoidal waves \cite{TakashimaVisualAtoms2023}, and stylized procedural textures \cite{SalehiStylized2025}, can provide surprisingly strong supervision signals for representation learning. These findings suggest that analytically generated patterns can capture fundamental visual statistics without relying on curated image datasets.

Motivated by these findings, we revisit texture-based augmentation from a procedural perspective. In particular, we observe that interference patterns naturally generate structured perturbations that span a wide range of spatial frequencies. Such perturbations resemble many types of real-world visual degradations and can be generated analytically with minimal computational cost. Among these patterns, Moire interference arises from the superposition of periodic structures and produces complex non-linear fringe patterns with rich spatial frequency variations.

Based on this insight, we propose \textbf{MoireMix}, a lightweight data augmentation framework that replaces external mixing image datasets with procedurally generated interference patterns. As illustrated in \cref{fig:abst}, instead of sampling textures from secondary image collections or computationally expensive generative models, MoireMix synthesizes Moire patterns on-the-fly using a closed-form mathematical formulation. The generated patterns are rendered directly in memory during training, blended with input images, and immediately discarded after use. This design enables a storage-free augmentation pipeline while providing a virtually unlimited stream of structured perturbations.

Our contributions are summarized as follows:

\begin{itemize}

\item \textbf{Procedural Interference Augmentation.}
We introduce a new perspective on robustness-oriented data augmentation that leverages analytically generated interference patterns. Unlike prior approaches that rely on stochastic noise, feature mixing, or external image datasets, our method uses procedurally generated Moire interference to produce structured perturbations spanning diverse spatial frequencies.

\item \textbf{MoireMix: On-the-fly Procedural Texture Generation.}
We propose MoireMix, an efficient augmentation framework that synthesizes interference textures directly in memory using a closed-form mathematical formulation. The generated patterns are mixed with training images during training and immediately discarded, enabling a storage-free augmentation pipeline with negligible computational overhead.

\item \textbf{Robustness Improvements Across Distribution Shifts.}
Through extensive experiments on ImageNet-1k with Vision Transformers, we demonstrate that MoireMix consistently improves robustness across multiple benchmarks, including ImageNet-C \cite{HendrycksCorruptions2019}, ImageNet-R, and adversarial attacks, outperforming standard augmentation baselines and existing approaches that do not rely on external mixing datasets.

\end{itemize}

    
    

\section{Related Work}
\label{sec:related}

\subsection{Data Augmentation for Robustness}
Improving robustness against distribution shifts and visual corruptions remains a central challenge in computer vision \cite{HendrycksCorruptions2019}. Data augmentation has emerged as one of the most effective approaches for enhancing model robustness. Early approaches such as MixUp \cite{ZhangMixup2018} and CutMix \cite{YunCutMix2019} improve generalization by interpolating images and labels or by replacing spatial regions between samples. AugMix \cite{HendrycksAugMix2020} further improves robustness by combining multiple stochastic image transformations while maintaining consistency constraints.

More recent approaches incorporate texture-based perturbations to expose models to a wider range of visual statistics. PixMix \cite{HendrycksPixMix2022} demonstrates that blending input images with diverse texture images significantly improves robustness across multiple distribution-shift benchmarks.
Subsequent works such as IPMix \cite{HuangIPMix2023} and DiffuseMix \cite{IslamDiffuseMix2024} further expand this paradigm by introducing additional texture sources, including feature visualizations and diffusion-generated images. These approaches show that injecting diverse texture patterns during training can effectively improve robustness.

However, existing texture-based augmentation methods fundamentally rely on external collections of mixing images. Such datasets must be curated and stored as secondary resources, which increases storage requirements and complicates dataset management in practical deployments.

\subsection{Legal and Ethical Concerns in Mixing Sets}
While these recent texture-based augmentation methods have undeniably driven substantial improvements in model robustness, their fundamental reliance on external mixing sets inherently introduces critical legal and ethical challenges for practical deployment. PixMix and IPMix utilize fractal art appropriated from web communities (\eg, DeviantArt); since copyright ownership resides with individual artists, the licensing status of these images is often ambiguous, creating a barrier to commercial utilization. Furthermore, the use of FVis not only incurs high computational costs for generation but also presents issues regarding its origin. Since FVis is generated from CNN models trained on ImageNet~\cite{RussakovskyImageNet2015}, it follows that FVis inherits these ethical and licensing constraints, making its commercial use inadvisable.

Similarly, DiffuseMix relies on large-scale diffusion models (\eg, Stable Diffusion \cite{RombachCVPR2022}), which are trained on massive datasets like LAION \cite{SchuhmannNeurIPS2022} consisting of image-text pairs scraped from the internet. Consequently, the legal and ethical issues associated with DiffuseMix directly inherit the unresolved copyright and transparency issues of its training data. Recent studies have explicitly highlighted these ethical gray areas, demonstrating that large-scale diffusion models can memorize and perfectly reproduce copyrighted or private training images without attribution \cite{CarliniUSENIX2023, SomepalliCVPR2023}. 

Given these factors, existing state-of-the-art augmentation techniques are often unsuitable for industrial applications from the perspective of transparency and rights management. This underscores the urgent need for a transparent and legally clean alternative.

\subsection{Robustness Benchmarks}
To evaluate the image classification accuracy and robustness of models, we utilize several standard benchmarks commonly adopted in this field:

\noindent
\textbf{ImageNet-validation}:
A dataset for evaluating standard image classification performance, consisting of 50,000 images from the ImageNet dataset that were not used during training.

\noindent
\textbf{ImageNet-C}:
An evaluation benchmark designed to measure robustness against common corruptions. It consists of 50,000 ImageNet validation images subjected to 15 types of algorithmic corruptions (\eg, noise, blur, weather) applied at five severity levels \cite{HendrycksCorruptions2019}. The performance is evaluated using mean accuracy (mean acc) across all corruption types and severities.

\noindent
\textbf{ImageNet-R}:
A benchmark consisting of approximately 30,000 images across 200 ImageNet-1k classes. These images feature artistic renditions distinct from standard photographs, such as art, cartoons, sketches, and embroidered motifs, testing the model's generalization capability to out-of-distribution data \cite{HendrycksManyFaces2021}.

\noindent
\textbf{ImageNet-PGD}:
An evaluation setting to measure robustness against Projected Gradient Descent (PGD) adversarial attacks \cite{MadryPGD2018}. Across all experiments, we consistently configure the attack under the $L_\infty$ norm constraint with 50 steps, a perturbation magnitude of $\epsilon = 1/255$, and a step size of $\alpha = 0.25/255$, utilizing a reduced batch size of 16 to manage the computational overhead. These strictly constrained perturbation parameters are specifically chosen because applying larger attacks to highly expressive models like Vision Transformers on ImageNet degrades accuracy to near zero, which would obscure meaningful performance differences between augmentation methods.

\noindent
\textbf{ImageNet-Moire}:
A custom benchmark simulating image degradation caused specifically by Moire patterns. We adopt the Moire generation pipeline proposed by Yuan \etal \cite{YuanDeMoire2019} for the AIM 2019 Challenge on Image Demoireing—which includes Liquid Crystal Display (LCD) pixel resampling, geometric transformation, Bayer Color Filter Array (CFA) simulation, signal processing, and JPEG compression—and apply it to the ImageNet validation set. We define this collection of 50,000 Moire-degraded images as the \textbf{ImageNet-Moire} benchmark.

\section{Proposed Method}
\label{sec:method}
This section introduces \textbf{MoireMix}, a robustness-oriented data augmentation framework based on procedurally generated interference patterns. The key idea is to replace external mixing image datasets with analytically generated textures that can be synthesized on-the-fly during training. These procedurally generated patterns are then integrated into a stochastic mixing procedure to create diverse perturbations during training. The overall workflow of the proposed method is summarized in Algorithm~\ref{alg:moiremix}.

\subsection{Motivation: Interference Patterns for Robustness}
Robustness-oriented data augmentation aims to expose models to diverse visual perturbations during training. Previous studies have shown that image classification models exhibit a strong bias toward texture cues~\cite{GeirhosTexture2019}. Consequently, modifying texture statistics during training can significantly improve generalization under distribution shifts.

A key requirement for effective texture perturbation is the ability to generate diverse spatial frequencies and complex edge structures. Simple procedural patterns such as stripes or sinusoidal waves contain only a single dominant orientation and frequency component, which limits their ability to produce sufficiently diverse perturbations.

Interference patterns provide a natural solution to this limitation. When periodic structures interact, their superposition generates non-linear fringe patterns that contain a rich spectrum of spatial frequencies. These patterns exhibit complex structural variations across orientations and scales, resembling many types of real-world visual degradations. Importantly, such interference patterns can be generated analytically with minimal computational cost.

Among various interference phenomena, \textbf{Moire patterns} are particularly suitable for augmentation because they naturally produce large-scale spatial variations while preserving fine-grained high-frequency structures. This property enables Moire patterns to generate perturbations that span multiple spatial scales, making them highly effective for robustness-oriented training.


\begin{algorithm}[tb]
\caption{On-the-fly MoireMix Data Augmentation}
\label{alg:moiremix}
\textbf{Input:} Training image $x \in \mathbb{R}^{W \times H \times 3}$, max components $N_{\text{max}} = 5$, max frequency $f_{\text{max}} = 100$, max iterations $k = 4$, beta parameter $\beta = 4$ \\
\textbf{Output:} Augmented image $x'$
\begin{algorithmic}[1]
\STATE \COMMENT{\textbf{Stage 1: On-the-fly Moire Generation}}
\STATE $N \sim \mathcal{U}\{1, N_{\text{max}}\}$
\STATE $M \gets \mathbf{0}^{W \times H}$
\FOR{$i = 1$ \TO $N$}
    \STATE $c_{x_i} \sim \mathcal{U}(0, W),\ c_{y_i} \sim \mathcal{U}(0, H)$ \COMMENT{Random center position}
    \STATE $f_i \sim \mathcal{U}(1, f_{\text{max}})$ \COMMENT{Random frequency}
    \FORALL{pixels $(u, v)$ in $W \times H$}
        \STATE $d'_i \gets \frac{1}{W}\sqrt{(u-c_{x_i})^2+(v-c_{y_i})^2}$
        \STATE $M(u, v) \gets M(u, v) + \sin(2\pi \cdot f_i \cdot d'_i)$
    \ENDFOR
\ENDFOR
\STATE $M \gets \text{MinMaxNormalize}(M, 0, 255)$
\STATE $x_{\text{moire}} \gets \text{RepeatChannels}(M, 3) / 255.0$ \COMMENT{Replaces external $x_{\text{mixing\_pic}}$}
\STATE
\STATE \COMMENT{\textbf{Stage 2: Standard PixMix Pipeline~\cite{HendrycksPixMix2022}}}
\STATE $x' \gets \text{RandomChoice}([\text{Augment}(x), x])$ \COMMENT{Initialize base image}
\STATE $rounds \sim \text{RandomChoice}([0, 1, \dots, k])$ \COMMENT{Random count of mixing rounds}
\FOR{$j = 1$ \TO $rounds$}
    \STATE $x_{\text{mix\_image}} \gets \text{RandomChoice}([\text{Augment}(x), x_{\text{moire}}])$
    \STATE $op \sim \text{RandomChoice}([\text{Add}, \text{Multiply}])$
    \STATE $x' \gets \text{MixOp}(x', x_{\text{mix\_image}}, \beta, op)$ \COMMENT{Apply Beta blending}
\ENDFOR
\STATE $x' \gets \text{Clip}(x', 0, 1)$
\STATE \textbf{return} $\text{Standardize}(x')$ \COMMENT{Apply ImageNet Normalization}
\end{algorithmic}
\end{algorithm}

\subsection{Procedural Moire Generation Algorithm}
The core of MoireMix is a procedural generator that synthesizes Moire interference patterns, detailed in Stage 1 of Algorithm \ref{alg:moiremix}. Let $M(u, v)$ denote the generated Moire image with spatial resolution $W \times H$. 

Each pattern is constructed by superimposing multiple radial wave components. The number of components $N$ is randomly sampled from a uniform distribution with an upper bound $N_{\text{max}} = 5$ (Algorithm \ref{alg:moiremix}, line 2). For each component $i$ ($1 \le i \le N$), we randomly sample a center position $c_i = (c_{x_i}, c_{y_i})$ and the frequency $f_i$(lines 5--6). The frequency range is set to $1 \le f_i \le 100$.

For a pixel location $p(u,v)$, the radial distance $d_i(p)$ to the component center $c_i$ is defined as
\begin{equation}
    d_i(p) = \sqrt{(u-c_{x_i})^2+(v-c_{y_i})^2}
\end{equation}
Using the normalized distance $d'_i(p) = d_i(p) / W$ (line 8), the pixel value for each component is determined by a sine wave and iteratively accumulated into the base image (line 9):
\begin{equation}
    M(u, v) \gets M(u, v) + \sin(2\pi \cdot f_i \cdot d'_i(p))
\end{equation}
The final Moire image $M(u, v)$ is obtained by accumulating all waves and applying Min-Max normalization to map values into the range $[0, 255]$ (line 12). 

To integrate this single-channel Moire pattern into the standard RGB training pipeline, we replicate its values across the three color channels to form $x_{\text{moire}}$ and normalize it to the $[0, 1]$ floating-point range (line 13). This formulation produces highly diverse interference patterns while maintaining extremely low computational cost.

\subsection{On-the-fly MoireMix Augmentation Pipeline}
\textbf{MoireMix} refers to our proposed dynamic augmentation pipeline that perfectly integrates into the standard PixMix framework \cite{HendrycksPixMix2022}. As illustrated in Stage 2 of Algorithm \ref{alg:moiremix}, instead of pre-generating and storing a static dataset to substitute FVis, MoireMix employs a strictly \textbf{on-the-fly} generation strategy. 

The procedurally generated patterns are integrated into a stochastic mixing-based augmentation pipeline. The augmented base image $x'$ is first initialized by randomly selecting either the original input image $x$ or its augmented version $\text{Augment}(x)$ (line 16). We then determine the number of mixing rounds by randomly sampling from $\{0, 1, \dots, k\}$ (line 17), where $k=4$ in our experiments.

In the mixing loop (lines 18--22), the target blending image $x_{\text{mix\_image}}$ is stochastically selected. Crucially, we replace the external mixing dataset entirely with our mathematically rendered $x_{\text{moire}}$ (line 19). A blending operation (\eg, addition or multiplication) is uniformly chosen (line 20), and the pattern is combined with the base image using a mixing weight sampled from a Beta distribution with $\beta=4$ (line 21). Finally, the mixed image is clipped to the valid $[0, 1]$ range and standardized using the ImageNet mean and standard deviation (lines 23--24).

This deliberate architectural choice provides crucial practical advantages. Because the Moire pattern $x_{\text{moire}}$ is mathematically rendered in memory for every single training iteration and immediately discarded, it completely eliminates external storage overhead. Supported by the extreme efficiency of our Moire formula (requiring approximately 0.0026 seconds per image), this generation seamlessly integrates into the training loop without introducing computational bottlenecks. More importantly, it consistently exposes the model to an infinitely diverse, unrepeated set of textures, preventing overfitting to a fixed mixing set. Furthermore, guaranteeing perfect reproducibility across different research groups is as simple as fixing the random seed.

\begin{table}[t]
\centering
\caption{Performance comparison of various data augmentation methods on ImageNet-1k using a ViT-Base model. We report standard accuracy (IN-val) alongside robustness metrics including ImageNet-C, ImageNet-R, PGD adversarial attacks, and Moire-degraded images. Bold values indicate the best results among comparable external-data-free methods.}
\label{tab:main_results}
{
\begin{tabular}{l|c|cccccc}
\toprule
Method & \shortstack{External\\Data-free} & IN-C & IN-R & IN-PGD & IN-Moire & IN-val  \\
\midrule
Baseline & \cmark & 34.4 & 15.3 & 1.3 & 20.3 & 69.2 \\
Cutout   & \cmark & 36.6 & 16.4 & 1.8 & 21.6 & 70.5 \\
AugMix   & \cmark & 39.1 & 17.2 & 1.1 & 27.7 & 70.7 \\
RandAug  & \cmark & 41.8 & 19.6 & 1.9 & 27.5 & 73.1 \\
MixUp    & \cmark & 45.1 & 19.8 & 1.3 & 35.7 & 74.1 \\
CutMix   & \cmark & 41.0 & 19.3 & 1.2 & 27.9 & \textbf{76.2} \\
MoireMix (Ours)  & \cmark & \textbf{50.5} & \textbf{23.1} & \textbf{3.1} & \textbf{36.2} & 73.1 \\
\midrule
PixMix   & \xmark & 53.7 & 25.7 & 3.8 & 53.4 & 75.6 \\
IPMix    & \xmark & 52.9 & 18.9 & 1.8 & 40.1 & 71.5 \\
DiffuseMix & \xmark & 59.3 & 16.9 & 1.0 & 31.9 & 69.8 \\
\bottomrule
\end{tabular}
}
\end{table}

\section{Experiments}
\label{sec:experiments}

This section evaluates the effectiveness of the proposed MoireMix augmentation through the training performance.

Our experiments aim to answer the following questions:

\begin{itemize}
\item Does MoireMix improve robustness compared to standard augmentation methods?
\item How does MoireMix compare with other procedurally generated texture perturbations?
\item How does MoireMix affect model sensitivity across spatial frequencies?
\item Does the proposed augmentation generalize across model architectures and datasets?
\end{itemize}

\subsection{Experimental Setup}

We train a Vision Transformer (ViT-Base) model on ImageNet-1k using the AdamW optimizer~\cite{LoshchilovAdamW2019} with an initial learning rate of $3\times10^{-3}$ and weight decay of $0.05$. Training is conducted for 100 epochs with a batch size of 256.

The baseline training pipeline uses standard augmentations including RandomResizedCrop and RandomHorizontalFlip. To evaluate robustness, we report performance on several commonly used benchmarks including ImageNet-C (corruption robustness), ImageNet-R (out-of-distribution generalization), and adversarial robustness measured using PGD attacks.

\subsection{Comparison with Existing Data Augmentation Methods}


\noindent
\textbf{Baselines and Comparisons:}
We compare MoireMix with widely used augmentation methods including Cutout \cite{DeVriesCutout2017}, AugMix \cite{HendrycksAugMix2020}, RandAug \cite{CubukRandAugment2020}, MixUp \cite{ZhangMixup2018}, and CutMix \cite{YunCutMix2019}. We also include texture-based mixing approaches such as PixMix~\cite{HendrycksPixMix2022}, IPMix~\cite{HuangIPMix2023}, and DiffuseMix~\cite{IslamDiffuseMix2024}, utilizing a subset of 4,700 FVis images for PixMix and the reported optimal configurations for IPMix and DiffuseMix.

\noindent
\textbf{Results:}
Table~\ref{tab:main_results} presents the results. Among augmentation methods that do not rely on external mixing datasets, MoireMix achieves the strongest overall robustness.
In particular, MoireMix improves ImageNet-C performance by 16.1 points over the Baseline and by 5.4 points over MixUp. Evaluations on ImageNet-R, ImageNet-PGD, and ImageNet-Moire similarly surpass standard baselines. While data-dependent methods like PixMix and DiffuseMix achieve higher scores on specific benchmarks (e.g., PGD), MoireMix provides highly competitive, comprehensive robustness improvements without external dependencies.

\begin{figure}[t]
\centering
\includegraphics[width=\textwidth]{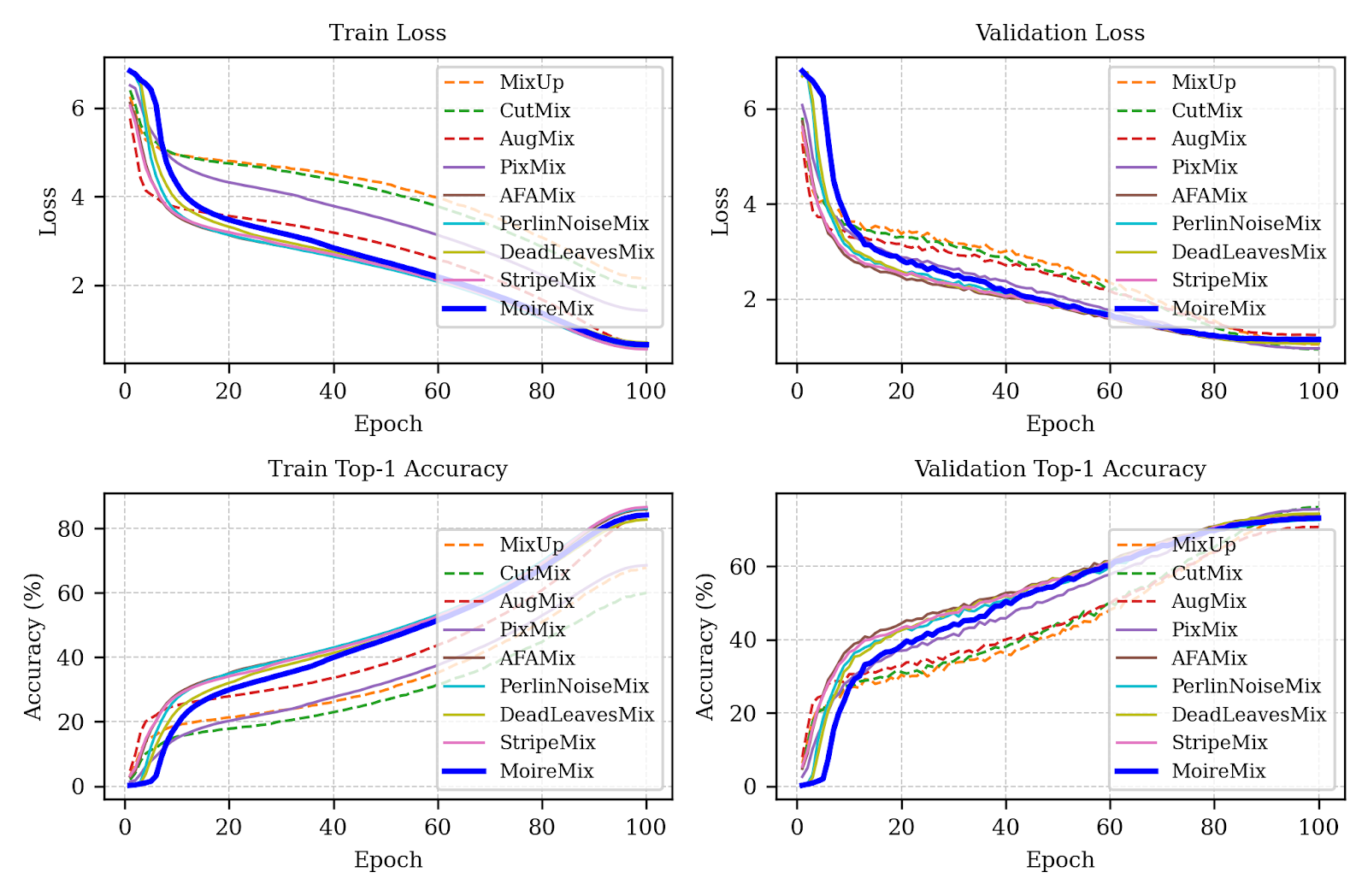}
\caption{Learning dynamics of the ViT-Base model on ImageNet-1k across 100 training epochs. We compare Train/Validation Loss and Top-1 Accuracy between traditional spatial augmentations (dashed lines) and texture-blending approaches (solid lines). MoireMix (thick blue line) provides a strong regularization effect, characterized by slower initial convergence but leading to a stable and robust final model state.}
\label{fig:learning_curve}
\end{figure}

\subsection{Analysis of Training Dynamics}
\label{sec:training_dynamics}
To understand the training behavior of MoireMix, we visualize the learning curves of the ViT-Base model over 100 epochs in Figure~\ref{fig:learning_curve}. Procedural texture-blending methods exhibit distinct training dynamics compared to traditional augmentations. Specifically, MoireMix imposes a heavy regularization penalty during early epochs, evident from its delayed initial convergence compared to PerlinNoiseMix or standard AugMix. While this strong penalty mitigates early memorization, it decreases early learning efficiency, necessitating sufficient training epochs for full convergence.

In later stages, MoireMix maintains a stable trajectory without severe divergence. However, a generalization gap between the final training accuracy ($\sim$85\%) and validation accuracy (73.1\%) remains, indicating that while the procedural noise provides effective regularization, partial memorization still occurs.

\begin{figure}[t]
\centering
\includegraphics[width=\textwidth]{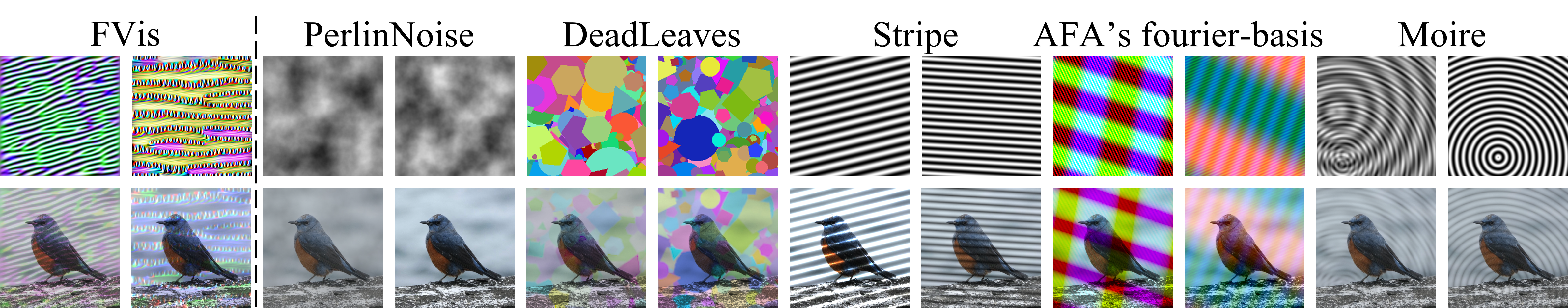}
\caption{Visual examples of procedurally generated texture images utilized as mixing sets in our on-the-fly augmentation pipeline. These specific examples illustrate the augmentation results obtained through a single addition operation. By applying the blending process once, the pipeline efficiently introduces diverse structural perturbations without external dependencies, storage overhead, or computational bottlenecks during training.}
\label{fig:generated_textures}
\end{figure}

\subsection{Comparison with Alternative Procedurally Generated Images}
To evaluate the specific impact of concentric interference fringes (Moire patterns), we integrated four alternative mathematically generated textures (Perlin Noise, Dead Leaves \cite{BaradadNoise2021}, Stripe, and AFA's Fourier-basis) into our on-the-fly pipeline under identical conditions.

\noindent
\textbf{Perlin Noise}: A gradient noise algorithm \cite{PerlinNoise1985} generating smooth, continuous variations dominated by low-frequency components (Figure~\ref{fig:generated_textures}a).

\noindent
\textbf{Dead Leaves \cite{BaradadNoise2021}}: Generates images mimicking natural occlusion statistics by stacking random geometric shapes, providing sharp edges and discrete structures (Figure~\ref{fig:generated_textures}b).

\noindent
\textbf{Stripe}: A simple periodic pattern based on sine waves with a single spatial frequency and direction, testing if complex interference is necessary.

\noindent
\textbf{AFA's Fourier-basis}: Adapted from \cite{VaishAFA2024}, assigning different frequency components to RGB channels to generate spectral images (Figure~\ref{fig:generated_textures}d).

\noindent
\textbf{Results:}
Superimposing any of these diverse textures surpasses the Baseline. Notably, Perlin Noise exhibited outstanding performance on ImageNet-C, indicating its effectiveness against corruptions where low-frequency components are preserved. Conversely, its performance against PGD attacks was significantly inferior. 

While StripeMix, AFAMix, and MoireMix demonstrate relative improvements against PGD attacks, absolute accuracy remains critically low (e.g., 3.1\% for MoireMix). This suggests that while structural texture blending offers some defensive benefits, models remain largely vulnerable to targeted adversarial perturbations. These findings imply that no single texture is universally optimal; the superimposing pattern should be designed based on the prioritized robustness type.

\begin{table}[t]
\centering
\caption{Ablation study comparing different procedural texture types within the on-the-fly mixing pipeline. We report the average CPU generation time per image in seconds alongside standard and robustness benchmarks. MoireMix provides a superior balance between computational efficiency and robustness across diverse corruption types, whereas methods like Perlin Noise introduce significantly higher latency during the training process.}
\label{tab:onthefly_results}
{
\begin{tabular}{l|c|ccccc}
\toprule
Method & Time [s] & IN-C & IN-R & IN-PGD & IN-Moire & IN-val \\
\midrule
Baseline       & -      & 34.4 & 15.3 & 1.3 & 20.3 & 69.2 \\
PerlinNoiseMix & 0.0592 & \textbf{54.9} & 20.6 & 1.2 & 32.1 & 72.6 \\
DeadLeavesMix  & 0.0050 & 53.8 & 23.5 & 2.1 & 44.0 & \textbf{74.4} \\
AFA-Mix        & 0.0172 & 48.6 & \textbf{24.0} & \textbf{4.7} & \textbf{45.9} & 74.2 \\
StripeMix      & 0.0009 & 50.4 & 23.0 & 2.9 & 35.9 & 73.4 \\
MoireMix (Ours)& 0.0026 & 50.5 & 23.1 & 3.1 & 36.2 & 73.1 \\
\bottomrule
\end{tabular}
}
\end{table}

\noindent
\textbf{Computational Efficiency vs. Robustness:} 
To assess practical viability, we measured the average CPU generation time per image (Table~\ref{tab:onthefly_results}). While Perlin Noise achieves excellent robustness against corruptions, its generation time (0.0592s) significantly slows down training. In contrast, MoireMix requires only 0.0026s per image. This exceptional efficiency ensures that data-loading does not become a bottleneck, making MoireMix a highly scalable and optimal solution for balancing robustness and training throughput.

\begin{figure}[t]
\centering
\includegraphics[width=\textwidth]{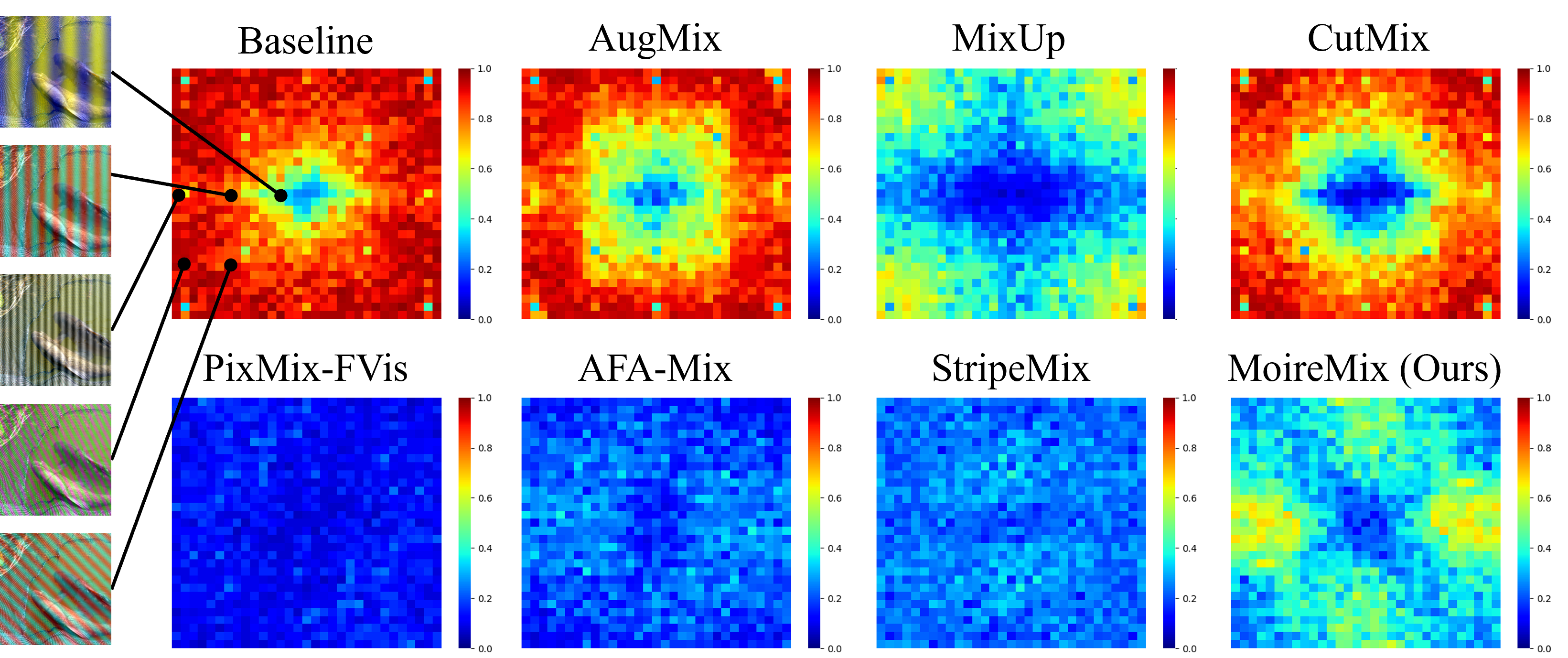}
\caption{Fourier sensitivity heatmaps visualizing robustness across spatial frequencies and orientations. Each pixel represents the classification error under specific Fourier basis noise. The center indicates low frequencies, while the periphery represents high frequencies. MoireMix exhibits a more generalized reduction in error across the spectrum compared to AFA-Mix and StripeMix, which show localized robustness likely due to evaluation bias.}
\label{fig:fourier_analysis}
\end{figure}

\subsection{Fourier Sensitivity Analysis}
Since Moire patterns inherently generate a rich spectrum of frequencies, we analyze how they influence model behavior in the frequency domain, following the methodology of Yin \etal \cite{YinFourier2019}. By injecting individual Fourier basis perturbations into test images, we evaluate the classification error rate to identify the specific frequency components against which the model has acquired robustness.

Figure~\ref{fig:fourier_analysis} displays these results. While AugMix and CutMix show minimal improvements over the Baseline, MixUp and MoireMix demonstrate a substantial reduction in overall error rates, confirming improved robustness across a broad frequency band. Although PixMix-FVis, AFA-Mix, and StripeMix achieve near-perfect classification against noise across all directions, this likely reflects an evaluation bias, as they superimpose textures during training that are structurally identical or highly similar to the evaluation noise. In contrast, MoireMix enhances robustness without relying on patterns that directly overlap with the evaluation noise, demonstrating a more generalized frequency-domain robustness.


\begin{table}[t]
\centering
\caption{Hyperparameter ablation for MoireMix generation, focusing on the number of superimposed waves and the sampled frequency range. Experiments were conducted using a ViT-Tiny model for 30 epochs. The results demonstrate that increasing structural complexity and utilizing high-frequency bands contribute significantly to generalization. Default settings used in main experiments are indicated by an asterisk (*).}
\label{tab:ablation_hyperparams}
\resizebox{\textwidth}{!}{
\begin{tabular}{l|cccccccccc|cccc}
\toprule
 & \multicolumn{10}{c|}{Number of Waves ($N$)} & \multicolumn{4}{c}{Frequency Range ($f_i$)} \\
\cmidrule(lr){2-11} \cmidrule(lr){12-15}
Metric & $1$ & $[1, 2]$ & $[1, 3]^*$ & $[1, 4]$ & $[1, 5]$ & $[1, 7]$ & $[1, 10]$ & $[1, 15]$ & $[1, 20]$ & $[1, 30]$ & \shortstack{Low \\ ($1 \le f_i \le 33$)} & \shortstack{Mid \\ ($34 \le f_i \le 66$)} & \shortstack{High \\ ($67 \le f_i \le 100$)} & \shortstack{Full$^*$ \\ ($1 \le f_i \le 100$)} \\
\midrule
IN-val & 60.3 & 59.8 & 60.5 & \textbf{62.0} & 60.7 & 61.1 & 61.8 & 61.3 & 61.8 & 61.7 & 60.3 & 60.1 & \textbf{62.0} & 60.5 \\
IN-R   & 12.1 & 11.7 & 11.8 & \textbf{13.0} & 12.2 & 12.4 & 12.8 & 12.7 & 12.7 & 12.7 & 11.0 & 12.5 & \textbf{12.8} & 11.8 \\
IN-C   & 33.0 & 32.3 & 32.7 & \textbf{34.3} & 32.6 & 33.5 & 33.4 & 33.4 & 33.6 & 33.0 & 31.0 & \textbf{32.9} & 32.8 & 32.7 \\
\bottomrule
\end{tabular}
}
\end{table}



\subsection{Ablation Study}
To understand the contribution of specific Moire pattern components, we conducted ablation studies on two key hyperparameters: the number of superimposed waves ($N$) and the frequency range ($f_i$), using a ViT-Tiny model trained for 30 epochs (Table~\ref{tab:ablation_hyperparams}).

\noindent
\textbf{Number of Waves ($N$):} Table~\ref{tab:ablation_hyperparams} (left) shows that increasing the interference complexity up to $N \in [1, 4]$ significantly improves performance across all metrics, yielding the best standard accuracy (IN-val), out-of-distribution generalization (IN-R), and robustness against common corruptions (IN-C). However, further increasing the number of waves (e.g., $N \ge 5$) leads to performance saturation or slight degradation. This is likely because excessively complex patterns lose their distinct structural characteristics and begin to resemble unstructured noise. Although $N \in [1, 4]$ achieved the highest overall scores, we adopted $N \in [1, 3]$ as the default setting for our main experiments to effectively balance structural complexity and generation speed.

\noindent
\textbf{Frequency Range ($f_i$):} Restricting sampled frequencies to specific bands (Table~\ref{tab:ablation_hyperparams}, right) reveals that high-frequency patterns ($67 \le f_i \le 100$) dramatically improve both standard accuracy (62.0\%) and robustness. This indicates that high-frequency non-linear interference is a primary driver of robustness improvements. We utilized the unconstrained \textit{Full} range ($1 \le f_i \le 100$) as a task-agnostic default, but these results highlight the potential for task-specific frequency tuning.

\begin{table}[t]
\centering
\caption{Robustness evaluation on CIFAR-10 and CIFAR-100 using WideResNet. Values represent Top-1 Accuracy (\%); higher is better.}
\label{tab:cifar_results}
{
\begin{tabular}{l|ccc|ccc}
\toprule
& \multicolumn{3}{c|}{CIFAR-10} & \multicolumn{3}{c}{CIFAR-100} \\
\cmidrule(lr){2-4} \cmidrule(lr){5-7}
Mixing Dataset & Clean & C & PGD & Clean & C & PGD \\
\midrule
Baseline             & 95.6 & 73.6 & 8.7 & 78.7 & 50.0 & 3.2 \\
Fractal arts         & 95.8 & 89.2 & 18.0 & 79.7 & 66.7 & 6.8 \\
FVis                 & 95.2 & 90.5 & 21.4 & 79.0 & \textbf{69.7} & \textbf{7.7} \\
FractalDB            & \textbf{96.0} & 88.1 & 7.8 & \textbf{80.0} & 65.0 & 1.5 \\
VisualAtom           & 95.6 & 89.2 & 6.1 & 78.6 & 66.6 & 1.5 \\
\midrule
\textbf{MoireDB (Ours, Static)} & 95.4 & \textbf{90.6} & \textbf{22.9} & 78.9 & 69.1 & 6.5 \\
\bottomrule
\end{tabular}
}
\end{table}

\subsection{Generalization to CNNs and Static Database Comparison}
To demonstrate that Moire-based augmentation generalizes beyond Vision Transformers and high-resolution ImageNet data, we conducted experiments using a Convolutional Neural Network (WideResNet-40-4~\cite{ZagoruykoWRN2016}) on CIFAR-10 and CIFAR-100~\cite{KrizhevskyCIFAR2009}. 

Adopting PixMix as the data augmentation pipeline, we compare a pre-generated set of Moire images (hereafter MoireDB) against Fractal arts and FVis. To ensure a fair comparison under identical conditions, we pre-generate an equivalent number of images for existing formula-driven datasets, namely FractalDB~\cite{KataokaIJCV2022} and VisualAtom~\cite{TakashimaVisualAtoms2023}.

\noindent
\textbf{Results:} Table~\ref{tab:cifar_results} presents the robustness evaluation on CIFAR-10 and CIFAR-100, including performance on clean data, common corruptions (CIFAR-10-C and CIFAR-100-C), and adversarial attacks (PGD). Even as a static database, MoireDB significantly outperforms the Baseline. On CIFAR-10, MoireDB achieves the best robustness scores, recording a corruption accuracy of 90.6\% and an adversarial accuracy of 22.9\%, effectively surpassing Fractal arts and FVis. On CIFAR-100, MoireDB remains highly competitive, outperforming mathematically generated datasets like FractalDB and VisualAtom, and performing closely to FVis.


\section{Discussion}
\label{sec}

Recent robustness-oriented augmentation methods often rely on external texture datasets such as fractal images or feature visualizations. In contrast, our results suggest that procedurally generated patterns can serve as an effective alternative to such external mixing datasets. By generating perturbations analytically, MoireMix provides an effectively unlimited space of texture variations without requiring curated image collections. This highlights a broader perspective for augmentation design, where structured perturbations can be synthesized directly from mathematical formulations rather than collected from external sources.

A key factor behind the effectiveness of MoireMix is the multi-scale frequency structure produced by interference patterns. The superposition of periodic structures generates spatially varying fringes containing both low- and high-frequency components. This enables the model to encounter perturbations spanning a broad frequency spectrum during training, which contributes to improved robustness across multiple benchmarks.

Despite these advantages, several limitations remain. While MoireMix improves robustness across corruption and distribution-shift benchmarks, it does not outperform state-of-the-art texture-based augmentation methods such as PixMix in all settings. In addition, procedural interference represents only one family of analytic perturbations, and other procedural generators may provide complementary robustness benefits. Exploring richer families of procedural augmentations is, therefore, an important direction for future research.

\section{Conclusion}
\label{sec:conclusion}

In this paper, we introduced \textbf{MoireMix}, a data augmentation method based on procedurally generated Moire interference patterns. By synthesizing mixing textures analytically, the proposed approach eliminates the need for external mixing image datasets while providing a diverse set of structured perturbations during training.

Through extensive experiments and frequency-domain analysis, we showed that the key to improving model robustness lies in the superimposition of diverse frequency components. We demonstrated that MoireMix effectively achieves this through the mathematical generation of interference fringes. Furthermore, our proposed method consistently improves model robustness over standard baselines in external-data-free settings, while strictly satisfying the practical requirements of avoiding external mixing datasets and remaining label-preserving.

Ultimately, these findings suggest that MoireMix aligns perfectly with the growing demand for sustainable, reproducible, and ethically responsible AI development, offering a highly practical approach for deploying robust models in real-world industrial applications.

\section*{Acknowledgement}
This work was supported by the AIST policy-based budget project ``R\&D on Generative AI Foundation Models for the Physical Domain.'' This work was supported by Japan Science and Technology Agency (JST) as part of Adopting Sustainable Partnerships for Innovative Research Ecosystem (ASPIRE), Grant Number JPMJAP2518. We used ABCI 3.0 provided by AIST and AIST Solutions with support from ``ABCI 3.0 Development Acceleration Use.''

\bibliographystyle{splncs04}
\bibliography{main}

\end{document}